# An Ensemble of Convolutional Neural Networks to Detect Foliar Diseases in Apple Plants

Kush Vora
Department of Computer Engineering
K.J. Somaiya College of Engineering
Mumbai, India
kush.vora@somaiya.edu

Dishant Padalia
Department of Electronics and Telecommunications
K.J. Somaiya College of Engineering
Mumbai, India
dishant.padalia@somaiya.edu

*Abstract*— Apple diseases, if not diagnosed early, can lead to massive resource loss and pose a serious threat to humans and animals who consume the infected apples. Hence, it is critical to diagnose these diseases early in order to manage plant health and minimize the risks associated with them. However, the conventional approach of monitoring plant diseases entails manual scouting and analyzing the features, texture, color, and shape of the plant leaves, resulting in delayed diagnosis and misjudgments. Our work proposes an ensembled system of Xception, InceptionResNet, and MobileNet architectures to detect 5 different types of apple plant diseases. The model has been trained on the publicly available Plant Pathology 2021 dataset and can classify multiple diseases in a given plant leaf. The system has achieved outstanding results in multi-class and multi-label classification and can be used in a real-time setting to monitor large apple plantations to aid the farmers manage their yields effectively.

*Keywords*— *Plant Pathology, Convolutional Neural Networks, Ensembling, Deep Learning*

## I. Introduction

Plant diseases can cause mild symptoms to cataclysmic disasters that damage large areas planted with food crops. There are approximately 19,000 fungi that may lay dormant but alive on both living and dead plant tissues and begin reproducing when the conditions are favourable. Furthermore, wind, water, soil, insects, and other invertebrates can swiftly spread fungal spores and infect entire cropland. These diseases primarily affect leaves, but some may also infect stems and fruit, resulting in not only a significant loss of yield and massive losses for farmers but also a severe danger to a nation's food security.

Apples are one of the most extensively grown fruits in the world. The disease has long been one of the primary causes of deteriorating apple quality and production, directly impacting the agricultural economy's development. Apple plants are prone to bacterial infections like fire blight, fungal and water mold infections such as scab or rust, viral infections like the flat apple disease, and physiological disorders like chlorosis, sunscald, and cork spot. Many of these infections develop over time and become apparent only when they spread throughout the tree. However, for the scope of this research study, five different types of apple diseases are considered: scab, frogeye leaf spots, rust, powdery mildew, and complex. Fig 1. showcases the different types of diseases in the dataset. Apple scab is a fungal disease that starts from the bottom of the leaves and spreads to the fruits. These scabs are responsible for total foliage loss during the mid-summer and can make the tree susceptible to other bacterial and fungal diseases. Frogeye leaf spots start as brown spots on the fruit's ends and expand into circles that can damage the entire apple. With time, the leaves grow small purple-brown spots like a frog's eye, which might spread to the tree's limbs and destroy it. Rust, which is caused by a complex fungus, is one of the most prevalent forms of infection. The rust disease, popularly known as Cedar apple rust, develops orange and yellow patches on the leaves and distorts the fruits. Powdery mildew is the most easily detected fungal infection and is characterized by the growth of a white substance on the underside of a leaf, resulting in stunted growth and leaves covered with black spots.

The apples produced by these diseased plants pose a significant threat to the humans and animals that consume them, leading to substantial losses of resources. Hence, detecting these plant diseases as soon as possible is imperative to manage plant health and minimize its risks effectively. The traditional approach to managing and monitoring plant health is manually scouting and checking the plant at regular intervals. However, this is not only expensive and time-consuming but also prone to misjudgment and can lead to a delayed diagnosis. Although several deep learning models have been employed to detect plant diseases for a variety of crops, such as tomatoes, potatoes, rice, corn, and cherries, there are several limitations such as variability in visual symptoms of the same disease between cultivars, variances in image capture conditions, variable image backgrounds, non-uniform image backdrops, and noisy images. To address these issues, we present an ensemble approach to disease classification in apple plants. The images in the dataset were augmented to mimic various real-life scenarios, and an ensemble of MobileNet, Xception, and InceptionResNet has achieved the best results on the testing dataset, with accuracy, precision, recall, and f1-score of 0.8731, 0.8905, 0.9100, and 0.9001, respectively.

## II. Literature Review

A number of CNN based approaches have been designed and implemented in the literature that have been successful in predicting various crop diseases. These algorithms have been trained and tested on different datasets with varying diseases. One such dataset was developed by P. Jiang et al. The dataset consisted of apple leaf images taken in a laboratory and an apple orchard. Several image processing techniques and augmentation methods were employed to increase the size of the dataset to 26,377 images. The images in the dataset were infected with the following diseases: Alternaria leaf spot, Brown spot, Mosaic, Gray spot, and Rust. The researchers also proposed a real-time detection model, the INAR-SSD. The architecture consisted of a pre-network and feature extraction and fusion structure. Inception models from the GoogLeNet replaced a few layers of the VGG-Net. The feature extraction and fusion structure were created using the Rainbow concatenation approach to boost feature fusion performance. The proposed INAR-SSD achieved 78.80% mAP, with a detection speed of 23.13 FPS on the custom developed dataset. S. Baranwal et al. used a subset of apple leaves from the Plant Village dataset to train a convolutional neural network (CNN) which achieved a net accuracy of 98.54%. Due to the data imbalance, S. Yu et al. used focal loss as the loss function. To focus learning on hard misclassified samples, focal loss adds a modulating term to the cross-entropy loss. It is a dynamically scaled cross-entropy loss, with the scaling factor approaching zero as confidence in the correct class grows.

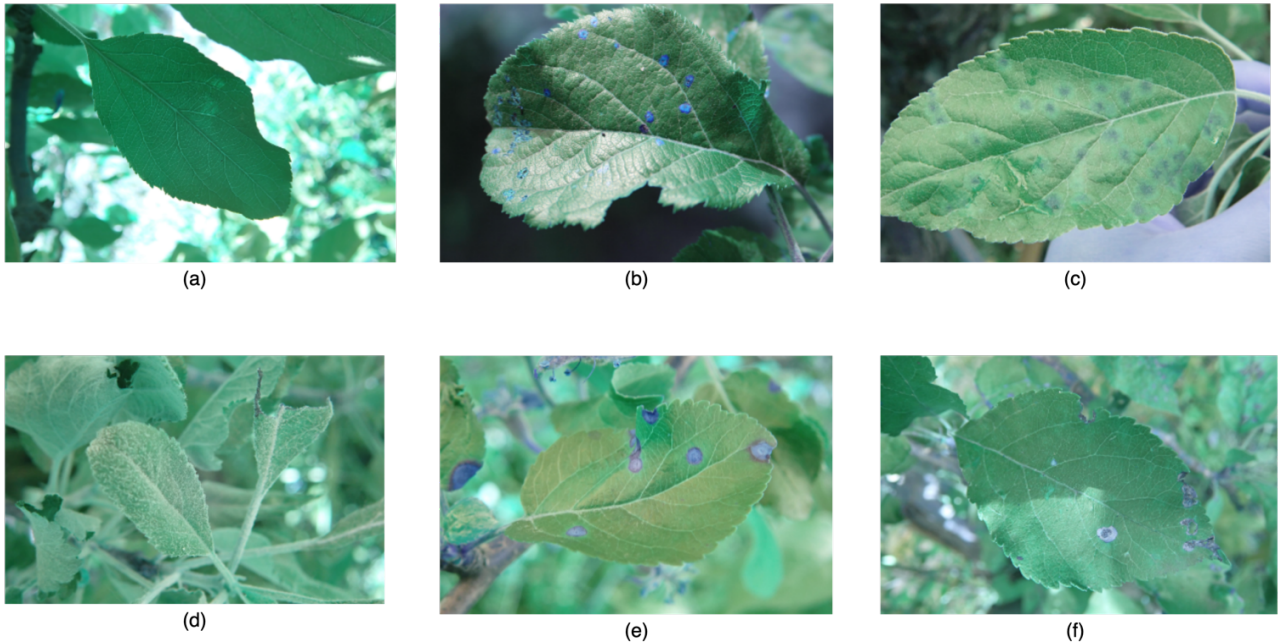

Fig. 1. Image samples of the Plant Pathology dataset used in the proposed work. (a) healthy apple leaf, (b) leaf infected with rust, (c) leaf infected with scab, (d) leaf infected with frogeye leaf spot, (e) leaf infected with powdery mildew, (f) leaf infected with complex diseases.

A number of approaches also involve ensembling CNN networks. S. Kejriwal et al. proposed an ensemble of three pre-trained convolutional neural networks (CNN) – ResNet101, Xception, and Inception-ResNet. The images were classified into six different classes, 5 of which belong to a type of disease and the last one being healthy. Several data augmentation techniques were employed to expand the Plant Pathology 2021 (FGVC8) dataset. The ensemble model achieved an F1 score of 96.25%, a precision of 97.43%, and a recall of 95.41%. P. Bansal et al. followed a similar ensemble approach. The three models ensembled were pre-trained DenseNet121, EfficientNetB7, and EfficientNet-NoisyStudent. The noisy student training approach is a semi-supervised learning strategy that uses a bigger or equal-sized student model and adds noise to the student during training to incorporate the ideas of self-training and distillation. The outputs from all three models were averaged to get the final prediction. The presented ensemble approach achieved an accuracy of 90% when classifying leaves with multiple images. A. Yadav et al. developed an Apple Foliar Disease Network (AFD-Net) model. In the model architecture, EfficientNetB3 and EfficientNetB4 are combined with a lambda layer followed by a dropout layer. The model was tested using the 5-fold cross-validation technique and achieved an accuracy of 98.70% on the plant pathology 2020 dataset and 92.60% on the plant pathology 2021 dataset.

Image augmentation is an essential preprocessing step when it comes to training a deep learning model. Besides basic augmentation techniques such as changes in the brightness, contrast, saturation, hue, and cropping. A. Yadav et al. employed two different augmentation techniques, viz. cut-mix, and mix-up, to augment the dataset and increase the model's learning capability. In the cut-mix algorithm, a part of the leaf is appended to a different image. This is performed to improve the localization capability of the model, whereas, in the mix-up algorithm, two samples from the dataset are blended by linear interpolation of their images and labels.

### III. MATERIAL

In the presented work, the Plant Pathology 2021 (FGVC8) dataset was used to train and evaluate the model. The dataset was built upon the existing Plant Pathology 2020 (FGVC7). The plant pathology 2020 dataset consists of 3642 RGB images with four classes: apple scab, cedar apple rust, complex and healthy. The pictures were taken from commercially produced cultivars in an unsprayed apple orchard at Cornell AgriTech during the 2019 growing season. A canon DSLR and mobile phones were used to capture the images in different lighting, angles, and with varying noise. Furthermore, images were captured during different times of the day, during different maturity stages of the plant, and under different focus settings. Images of plants showcasing multiple diseases were also captured to increase the complexities of the dataset. In 2021, the plant pathology 2021 dataset was released, which consisted of 18632 images with two added diseases, viz. powdery mildew and frogeye leaf spot, taking the total count of diseases in the dataset to 5. The dataset supports multiclass and multi-label classification, with 17,277 images having only 1 type of disease and 1,355 images having multiple diseases. The dataset distribution for both, multi-class and multi-label classification can be seen in Fig 2.

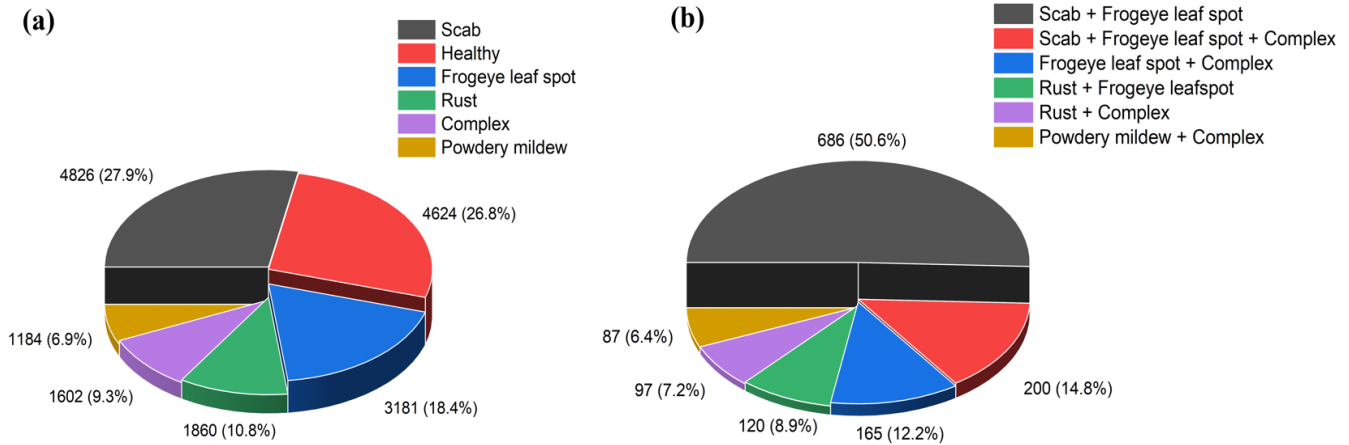

Fig. 2. Dataset distribution of the Plant Pathology 2021 dataset. (a) multi-class distribution of the dataset, (b) multi-label distribution of the dataset

## IV. DATA PREPARATION

The images in the Plant pathology dataset were of different sizes. To maintain uniformity, the images were resized to 256x256 before feeding them to the model. Post resizing, the images in the dataset were split in an 80-20 ratio. 14,906 images were used for training the presented architectures, and 3,726 images were used to evaluate the performance of the models. All the images in the training set were subjected to different augmentation techniques, viz. rotation, flipping, horizontal and vertical translation, and zooming. Augmentation techniques like shearing were applied to mimic real-world settings. Shearing distorts an image along an axis, changing the perception angle, allowing the model to see a particular image from different angles.

## V. METHODOLOGY

Deep learning has grown in popularity recently due to its ability to learn and comprehend computationally complex problems. Deep learning is a subtype of machine learning in which machines are trained to execute things that people are born with. Deep learning models may create new features independently, whereas traditional machine learning techniques rely on humans to discern between features. It has found success in a wide range of fields, including object recognition, text processing, audio processing, time series analysis, and gaming. Convolutional Neural Networks or ConvNets are one of the most widely used deep learning techniques. ConvNets can interpret pictures, assign relevance to items in the image using learnable weights and biases, and differentiate across images. The algorithm of a CNN is based on the model of a human visual cortex. CNNs offer numerous advantages over traditional multi-layer perceptrons, including the fact that they do not require manual feature engineering and are computationally much more efficient than conventional neural networks due to parameter sharing and dimensionality reduction. Some of the common CNN architectures that have achieved state-of-the-art results for several use cases include VGG, ResNet, DenseNet, InceptionNets, and EfficientNets. In the presented work, five different architectures, viz. ResNet, Xception, InceptionResNet, MobileNet, and NASNetMobile were used as a backbone, and additional layers were added to these base models for the specific task.

The model's performance was seen to degrade with particularly deep CNN architectures. The vanishing gradients were one of the causes behind this. The gradients from which the loss function is constructed decrease to zero after successive applications of the chain rule. As a result, the weights' values are never updated; hence, no learning happens. As a result, Kaiming He et al. introduced the ResNet design, which included skip connections (also called residual connections). These skip connections allowed gradients to flow from subsequent layers to the original filters, resolving the issue of disappearing gradients. ResNet designs were exceptionally deep convolutional neural networks that outperformed the ImageNet dataset. This work employed a 50-layer deep resnet (Resnet50) as the base structure and was trained for a total of 28 epochs.

The MobileNet architecture, designed for small-scale applications such as mobile apps, was open-sourced by Google Developers. These CNN architectures use depth-wise separable convolutions instead of standard convolutions. As a result, the number of parameters in the MobileNet model is relatively low compared to other designs. A depthwise separable convolution splits a kernel into two different kernels that execute depthwise and pointwise convolutions. In depth-wise separable convolutions, the depth dimensions are separated from the height and width dimensions, and then a 1x1 filter is used to cover the depth dimension, reducing the overall number of parameters. In our work, the MobileNet architecture was enhanced for the particular task by plugging different layers on top of the architecture. The architecture was trained for a total of 34 epochs, after which the model performance started to overfit on training data.

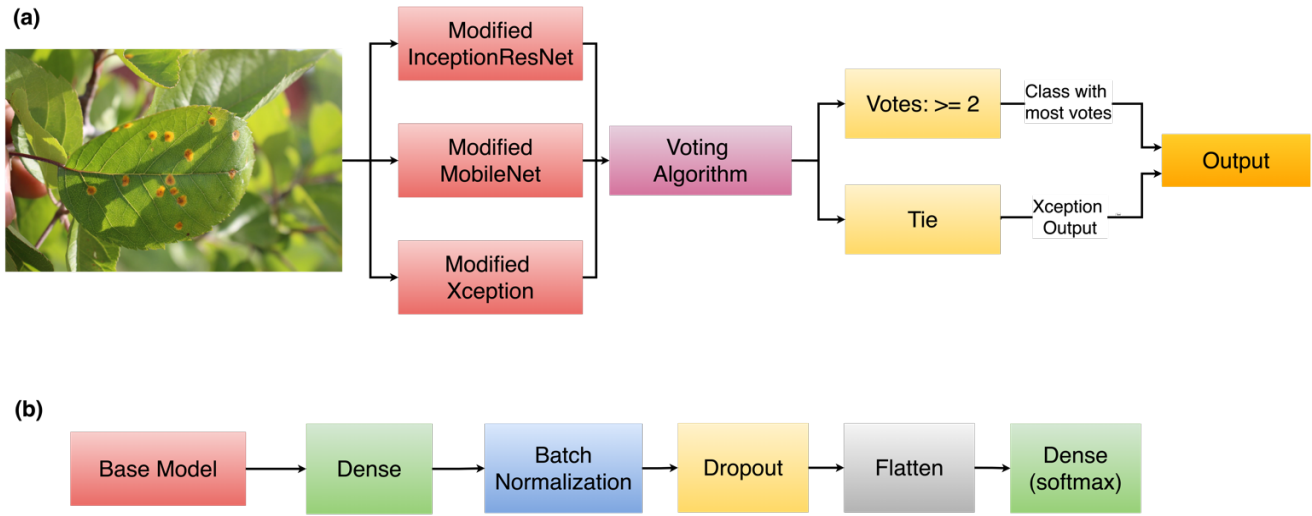

Fig. 3. Flow diagram of the proposed system. (a) A leaf image is passed through three modified architectures viz. Xception, InceptionResNet, and MobileNet. The output from each of the three architectures is then passed through a voting classifier to predict the final disease(s), (b) extra layers added to all the architectures.

In 2016, François Chollet presented the Xception architecture in association with Google Research. Xception is a deep convolutional neural network architecture that employs Depthwise Separable Convolutions. Xception stands for "Extreme Inception," and C. Szegedy's Inception architecture serves as the foundation of the Xception framework. Inception compresses the input pictures using convolutional layers before applying filters to each depth space. However, Xception reverses this procedure, applying the filters first and then compressing the image using a 1x1 convolution operation. Xception is based on two primary factors: Depthwise Separable Convolution, which is far more efficient than traditional convolutions, and shortcuts between convolutional layers, which are employed in ResNet design. Four layers were added on top of the Xception net to classify apple diseases. The performance of the network showed no improvement after 14 epochs.

Following the success of the Inception network and ResNets, C. Szegedy et al. proposed the InceptionResNet to reduce the complexity of InceptionV3. They demonstrated that training using residual connections significantly improved the training of Inception networks. This design's inception blocks were computationally less expensive than standard Inception blocks. Following the inception blocks was a 1x1 convolution known as the filter expansion, which increased the dimensionality of the filter bank to match the depth of the following layer. Furthermore, residual skip connections replaced the pooling operations within the Inception blocks. InceptionResNet took the shortest time to train on the Plant Pathology dataset. The model's performance was stagnant after 6 epochs.

In 2017, Barret Zoph et al. from Google Brain proposed the NASNet architecture. NASNet is an abbreviation for Neural Architecture Search Network. The NASNet uses reinforcement learning to choose the most suitable CNN architecture. It works on the premise of searching through a space of neural network configurations for the optimal combination of parameters such as filter sizes, output channels, strides, number of layers, and so on. NASNet's overall process consists of three major components: Search Space, Performance Estimation Strategy, and Search Strategy. Search Space defines the space of all possible neural networks for the specific task, Performance Estimation evaluates a potential neural network's performance based on its architecture rather than training or constructing it, and Search Strategy aids in identifying the best architecture using a variety of techniques such as random search, grid search, gradient-based strategies, evolutionary algorithms, and so forth. In the presented work, NASNetMobile was used, which has a relatively lesser number of trainable parameters when compared to the original NASNet architecture. The architecture achieved peak-performance after 31 epochs.

In the presented work, the above-mentioned architectures were modified for apple disease detection by adding different layers on top of the architectures. A combination of the dense layers, a batch normalization layer and the dropout layer performed best on the dataset. A dense layer was added, followed by batch normalization and a dropout layer with a dropout rate of 20%. Batch normalization was used for training very deep neural networks and to make sure that the contributions to a layer were normalized for each mini-batch. This helped us achieve the effect of settling the learning process and significantly reducing the number of training epochs needed to train deep neural networks. Dropout layer was added to avoid overfitting by ignoring randomly chosen neurons during training. Finally, an additional dense layer was added with an activation function of the sigmoid. For training all the models presented in this work; Adam (Adaptive Moment Estimation) optimizer was used with a learning rate of 0.0001. Adam optimizer accelerates the gradient descent algorithm and integrates the finest features of the AdaGrad and RMSProp methods to devise an optimization algorithm capable of dealing with sparse gradients on noisy issues. Binary cross entropy was used as a loss function to map the distance between the input class and the output prediction and evaluate the model's performance and a batch size of 32 was used to train the architectures.

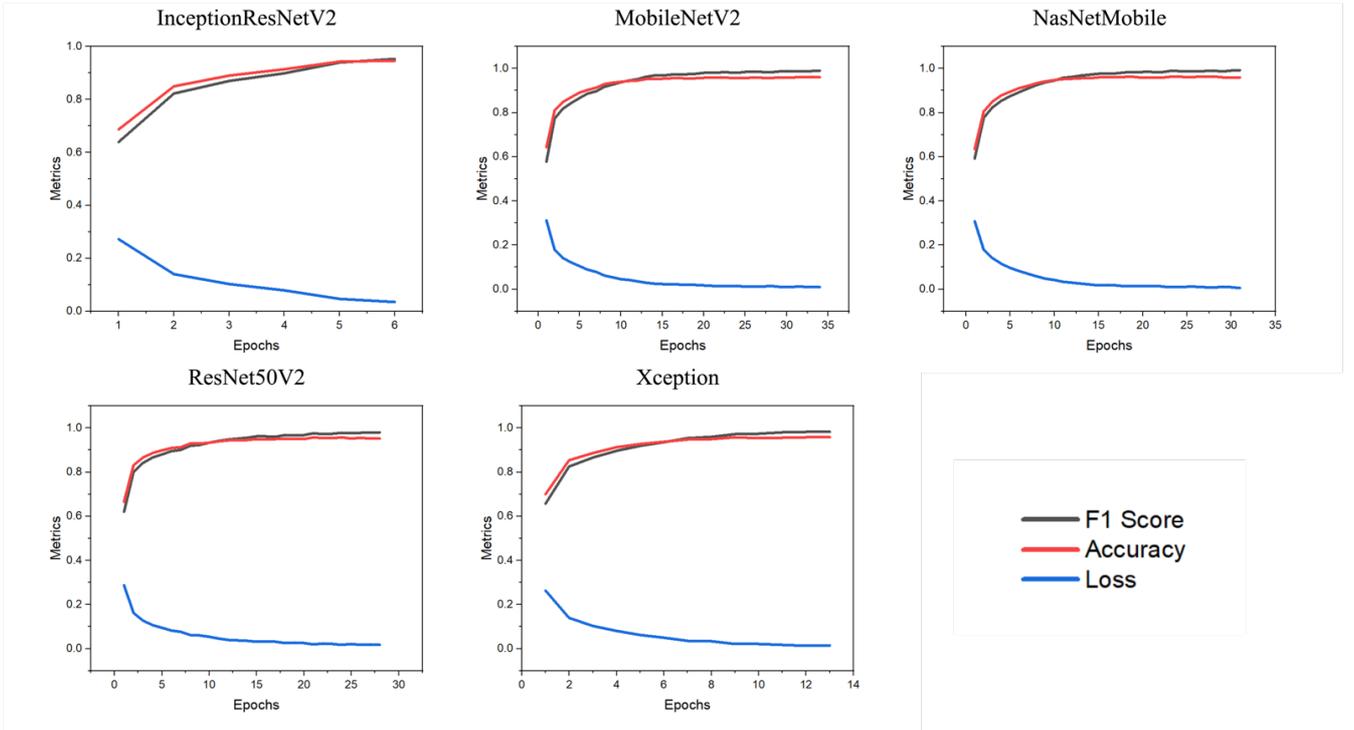

Fig. 4. Training accuracy, F1-score and loss for the architectures trained in the proposed work.

In general, neural networks function well for a specific class or feature while performing less accurately for some features that the model is unable to capture correctly; this is referred to as neural networks with a high variance. the predictions of several models trained on the same data as a voting system and selecting the output based on the class that obtains the most support or votes is a suitable method for reducing variance; this technique is known as ensembling. In other words, it is a process of merging numerous algorithms to improve their overall performance. In this study, we have trained five different state-of-the-art architectures and have ensembled all possible combinations of the same. An ensemble of MobileNet, Xception, and InceptionResNet performed the best on the testing data. According to the rules for our ensemble model, if the majority of the models vote for a given class, then it is considered the output prediction. In the event of a tie, the output of the Xception network is used as the prediction. This is because the Xception network showcased the best individual performance on the testing set with an F1 score of 0.8802. The entire flow structure of the proposed ensemble system can be seen in Fig 3.

## VI. RESULTS AND DISCUSSION

The trained models were assessed using a large testing set consisting of 3,726 images. Accuracy, precision, recall and F1 score were used as the performance evaluation metrics. The training performance (loss, accuracy, and F1 score) per epoch of the different models employed in the work can be seen in Fig 4. Table 1. shows a quantitative comparison of the different architectures developed in the presented work. The Xception network showcases the best individual performance on the dataset, with an F1 score of 0.88. Due to the smaller number of trainable parameters, the NasNetMobile architecture took only 270secs/epoch to train. Different combinations of the models were ensembled of which Xception, InceptionResNet and MobileNet performed the best on the dataset by achieving an accuracy, precision, recall, and F1-score of 0.8731, 0.8905, 0.9100, and 0.9001, respectively. Due to a lack of a standardized testing dataset, the existing proposed systems in the literature have worked on different subsets of the dataset or have trained their architectures on different types of diseases. Hence it is not possible to compare the presented work with others in the literature.

## VII. CONCLUSION AND FUTURE SCOPE

Agricultural crops such as fruits and vegetables provide food, oil, and fiber for consumption and are a major component of a country's economy. Hence, crop losses due to various bacterial infections can cause large sums of money to the farmers and the nation. Crop monitoring helps in detecting areas infested by disease-causing bacteria or fungus and allows for early treatment, significantly increasing disease control efficacy. However, the conventional approach of identifying illnesses by physically evaluating certain features of leaves, such as texture, color, and shape, is not always effective and calls for an automated approach which makes things easy for the farmers. In the presented work, we propose an ensemble approach to detect diseases in apple orchards. The proposed architecture can detect multiple diseases infecting a given plant and can be used in a real-time setting to monitor large fields of apple orchards. The developed ensemble of Xception, InceptionResNet, and MobileNet can also be extrapolated and trained for a wide variety of crops and be used to detect different diseases in the area. In addition, an automated monitoring system can be built on top of the model, which alerts the farmers when a plant infestation is detected.

TABLE 1. MODEL COMPARISON OF DIFFERENT ARCHITECTURES TRAINED IN THE PROPOSED WORK. THE TRAINED NETWORK OF XCEPTION PERFORMS THE BEST INDIVIDUALLY ON THE TESTING SET WHEREAS THE ENSEMBLE OF XCEPTION, INCEPTION-RESNET, AND MOBILENET OUTPERFORMS OTHER ENSEMBLE ARCHITECTURES.

| Model | Accuracy | Precision | Recall | F1 Score |
|---|---|---|---|---|
| ResNet50 | 0.8062 | 0.8278 | 0.8873 | 0.8552 |
| Xception | 0.8513 | 0.8674 | 0.8940 | 0.8802 |
| InceptionResNet | 0.8253 | 0.8454 | 0.9048 | 0.8737 |
| MobileNet | 0.8492 | 0.8708 | 0.8847 | 0.8773 |
| NASNetMobile | 0.8247 | 0.8489 | 0.8799 | 0.8622 |
| Xception + InceptionResNet + NasNet | 0.8684 | 0.8873 | 0.9080 | 0.8972 |
| MobileNet + Xception + ResNet | 0.8677 | 0.8901 | 0.9026 | 0.8962 |
| MobileNet + InceptionResNet + NasNet | 0.8693 | 0.8882 | 0.9062 | 0.8967 |
| MobileNet + InceptionResNet + ResNet | 0.8629 | 0.8848 | 0.9069 | 0.8956 |
| MobileNet + NasNet + ResNet | 0.8680 | 0.8894 | 0.9016 | 0.8951 |
| NasNet + Xception + ResNet | 0.8650 | 0.8854 | 0.9018 | 0.8934 |
| **MobileNet + Xception + InceptionResNet** | **0.8731** | **0.8905** | **0.9100** | **0.9001** |